# Generative method for aerodynamic optimization based on classifier-free guided denoising diffusion probabilistic model

Shisong Deng[1,2], Qiang Zhang[1,2], Zhengyang Cai[1,2]*

[1] School of Management, Hefei University of Technology, Hefei, China

[2] Key Laboratory of Process Optimization and Intelligent Decision-making, Ministry of Education, Hefei, China

*Corresponding author. E-mail: zycai@hfut.edu.cn;

Contributing authors: shisongdeng@mail.hfut.edu.cn; qiang_zhang@hfut.edu.cn;

**Abstract**

Recently, inverse design approach, which directly generates optimal aerodynamic shape with neural network models to meet designated performance targets, has drawn enormous attention. However, the current state-of-the-art inverse design approach for airfoils, which is based on generative adversarial network, demonstrates insufficient precision in its generating and training processes and struggles to reveal the coupling relationship among specified performance indicators. To address these issues, the airfoil inverse design framework based on the classifier-free guided denoising diffusion probabilistic model (CDDPM) is proposed innovatively in this paper. First, the CDDPM can effectively capture the correlations among specific performance indicators and, by adjusting the classifier-free guide coefficient, generate corresponding upper and lower surface pressure coefficient distributions based on designated pressure features. These distributions are then accurately translated into airfoil geometries through a mapping model. Experimental results using classical transonic airfoils as examples show that the inverse design based on CDDPM can generate a variety of pressure coefficient distributions, which enriches the diversity of design results. Compared with current state-of-the-art Wasserstein generative adversarial network methods,

CDDPM achieves a 33.6% precision improvement in airfoil generating tasks. Moreover, a practical method to readjust each performance indicator value is proposed based on global optimization algorithm in conjunction with active learning strategy, aiming to provide rational value combination of performance indicators (i.e. meeting aerodynamic constraints) for the inverse design framework. This work is not only suitable for the airfoils design, but also has the capability to apply to optimization process of general product parts targeting selected performance indicators.

**1 Introduction**

The rapid development of aviation technology has led to increasingly higher performance demands for aircraft, necessitating continuous pursuit of aerodynamic shape optimization to enhance aircraft speed, fuel efficiency, etc (Coder and Maughmer 2014). The aerodynamic shape of the airfoil is one of the important factors affecting aircraft performance, thus its optimization design is essential. This involves complex fluid mechanics and structural mechanics issues that require consideration of the interaction between multiple parameters (Li et al. 2023). Additionally, the design process also needs to take into account the balance between multiple performance indicators. For example, increasing lift will increase drag, and reducing drag may affect stability. However, traditional airfoil aerodynamic shape optimization design is typically based on trial-and-error experience. The optimal solution for engineering design is gradually found through iterative cycles of 'design modification - simulation experiment – redesign'. The simulation involved in this process consume significant time and computational resources, making it challenging to improve design efficiency. Fortunately, the emergence of various surrogate models has accelerated this process (Zuo et al. 2023; Yetkin et al. 2024). This has motivated researchers to delve deeper into exploring and exploiting the potential of airfoils. Despite the progress that has been made, it has not changed the way designers need to gradually search for target airfoils from a wide design space (Silva et al. 2021; Kilimtzidis and Kostopoulos 2023). The inverse design, however, aims to start from designated aerodynamic performance targets and directly generate the optimal shape design, which not only meets specific performance requirements, but also can reduce calculation complexity, providing designers with more intuitive solutions. Therefore, the key to current airfoil aerodynamic shape design optimization is how to conduct inverse design accurately and efficiently, enabling designers to swiftly identify the target airfoil shape.

Early airfoil inverse design primarily relied on optimization algorithms and decomposition techniques, such as orthogonal decomposition (Bui-Thanh et al. 2004), genetic algorithms (Wu et al. 2023; Shen et al. 2024), and improved optimization algorithm (Yetkin et al. 2023; Rastgoo and Khajavi 2023). The main idea of these methods is to search the geometric parameters of the airfoil through optimization algorithms to achieve predefined aerodynamic performance targets. While these methods were beneficial in the early stages of airfoil inverse design, they exhibit certain limitations, including a limited capacity to handle highly nonlinear problems, a propensity for falling into local optima, and sensitivity to initial values. With the advancement of deep learning technologies, data-driven design methods have been introduced to the field of airfoil inverse design with novel approaches and solutions.

In the initial stages, scholars such as Sun (Sun et al. 2015) employed Artificial Neural Networks (ANN) for the direct inverse design of airfoil/wings that meet aerodynamic characteristics. Similarly, Sekar et al. (Sekar et al. 2019) used Convolutional Neural Network (CNN) to derive airfoil shapes from pressure distribution coefficients. Both ANN and CNN fundamentally aim to establish a mapping relationship from aerodynamic performance features to airfoil shapes. This is exactly the reverse operation of the surrogate models in traditional design, which mapping geometry to corresponding flow conditions (Wu et al. 2020; Catalani et al. 2023). However, for the inverse design of airfoils, especially when considering performance features such as continuous pressure distribution coefficients corresponding to the airfoil itself, the selection and generation of airfoils are quite complex and challenging tasks for designers. Recently, the development of generative models has opened up new possibilities for inverse design (Gm et al. 2020). The advantage of this approach lies on its ability to learn from existing data, capturing complex aerodynamic characteristics, and introducing additional information to influence the generation of airfoil shapes. Conditional generation enables designers to have more precise control over the generation process to meet specific application scenarios or performance standards.

Currently, the field of airfoil inverse design predominantly focuses on the use of generative models such as generative adversarial networks (GAN) (Lei et al. 2021; Deng and Yi 2023), Variational Autoencoders (VAE) (Yonekura et al. 2022b; Wang et al. 2022; Yang et al. 2023b), and their enhanced variants (Sun et al. 2023; Xie et al. 2024; Liu et al. 2024). Researchers initially utilized lift, drag, and pitching moment coefficients as additional conditions to aid in airfoil generation (Yilmaz and German 2020; Yonekura and Suzuki 2021). Achour et al. (Achour et al. 2020) used CGAN to generate the corresponding airfoil based on the pre-calculated lift-to-drag ratio and shape area. However, due to the inaccuracy of some trailing edge shapes, 10% of the airfoils were difficult to converge. In order to generate a smoother airfoil, considering the difficult convergence characteristics of GAN itself, Yonekura et al. (Yonekura et al. 2022a) used the improved WGAN-GP model to generate a smooth airfoil based on the lift coefficient without any smoothing method. However, in practical design processes, the design focus for airfoils typically does not seek the airfoil with the highest lift-to-drag ratio. Instead, the focus is on stability considerations, and specific characteristics such as pressure distribution coefficient, drag divergence Mach number, etc. (Li et al. 2018; Wang et al. 2025) are selected as performance indicators. For example, a three-stage design based on WGAN was proposed by Lei (Lei et al. 2021) and Deng et al (Deng and Yi 2023), using the pressure feature as a specified condition for generating the corresponding airfoil. Furthermore, Wang (Wang et al. 2022) also integrated Conditional Variational Autoencoders (CVAE) and WGAN to map the Mach number distribution of specific characteristics to the airfoil and measure its smoothness.

The aforementioned studies have demonstrated substantial progress made by generative models (e.g., GAN and VAE) in airfoil inverse design, but there are still three obvious defects that can be further improved: (1) Most GAN- and VAE-based design methods exhibit significant instability during training and suffer from incomplete capture of aerodynamic features. (2) Performance indicators conforming to physical flow laws often rely on manual design experience, and defining such indicators typically requires extensive experimental work to validate their effectiveness. Although some studies have attempted to use the performance indicators as controllable variables in GAN or VAE through algorithms, these indicators often function as intermediate variables within the model, and thus it is difficult to provide effective and interpretable feedback, which limits the model's overall control over the design performance. (3) Airfoil aerodynamic shape optimization involves multiple objectives, and multiple performance indicators need to be weighed in the design process. However, the existing research often focuses on improving the inverse design accuracy of the performance indicators, and lacks in-depth research on the coupling relationship between the performance indicators. How to determine target values for each performance indicator to ensure that the combination of indicators is self-consistent remains a challenging problem.

To address the above defects, this study proposes an airfoil inverse design method based on a classifier-free guided denoising diffusion probability model (CDDPM). The method consists of three main steps: first, the transonic pressure coefficient (CP) distributions are captured through the diffusion model according to the six pressure performance features such as the suction peak and pressure gradient. The diffusion model provides enhanced physical interpretability, facilitating better integration with the requirements of aerodynamic performance (Li et al. 2018; Yang et al. 2023a). Next, the mapping model is utilized to extract the physical information from the CP distribution and accurately reconstruct the corresponding airfoil geometry, ensuring that the generated airfoil adheres to the specified pressure features. By employing curve fitting techniques, the generated airfoils can be further applied to practical engineering designs. Finally, the performance features and coupling of the generated airfoils are verified, and the global optimization algorithm (EGO) is used to readjust each performance indicator in order to provide a combination of indicators that satisfy the aerodynamic constraints. The framework is also updated with an active learning strategy to improve the design efficiency and performance.

The primary contributions of this study are as follows:

1) A generalized airfoil reverse design process and data preparation procedure are summarized, which can accurately generate target airfoils based on specified performance features and can be directly extended to other inverse design fields under similar complex conditions.

2) An airfoil inverse design framework based on a classifier-free guided denoising diffusion probability model is developed, capable of directly capturing the aerodynamic features based on targeted aerodynamic performance indicators, thereby generating optimal airfoil shapes. Compared with the current state-of-the-art methods, the proposed framework achieves approximately a 33.6% improvement in design accuracy.

3) Through the study of performance indicators, the interactions between different design objectives in the inverse design process are revealed. Moreover, a practical method to readjust each performance indicator value is proposed based on EGO in conjunction with active learning strategy, which not only realizes the updating of the inverse design framework, but also addresses the challenging problem of selecting performance indicators that comply with aerodynamic constraints.

This paper is organized as follows: Section 2 presents the problem definition and research framework for airfoil inverse design. The diffusion model, mapping model, and optimization model in the CDDPM-based airfoil inverse design framework are then presented in Section 3. The construction of the dataset and the performance analysis of the models in the framework are presented in Section 4. Section 5 verifies the superiority of the proposed methodology through comparative experiments and conducts a coupling analysis of the performance metrics to provide a more reasonable set of performance indicators. Finally, Section 6 concludes this study with suggestions for future work.

**2 Problem statement**

Inverse problems are central to many scientific and engineering disciplines, involving complex computational processes to derive inputs from outputs. Unlike straightforward forward problems, inverse problems often exhibit pathological characteristics, meaning solutions may not exist, may not be unique, or may be highly sensitive to small variations in input data. The inverse design of an airfoil is based on specified aerodynamic performance indicators, which are inverted to deduce the corresponding airfoil geometry. Mathematically, this design process can be formulated as an optimization problem, with the objective of finding an airfoil geometry $A$ that minimizes the difference between the corresponding aerodynamic performance indicators $P(A)$ and the target performance indicators $P_{target}$. This problem can be formalized as:

$$\min_{A} \|P(A) - P_{target}\|. \tag{1}$$

In this framework, $P_{target}$ usually involves multiple mutually coupled parameters, such as peak suction and pressure gradient, and there are complex nonlinear relationships between these indicators and the airfoil geometry, which increases the complexity of searching for solutions in the design space.

To help designers design target airfoils with specified performance features more accurately and efficiently, this paper proposes an airfoil inverse design method based on a classifier-free guided denoising diffusion probability model. This method is suitable for the early design stage of airfoils and can accurately inverse design the corresponding geometric airfoils based on target performances. It reduces the design search space and provides effective shape references and theoretical guidance for airfoil design. Figure 1 shows the flow of the inverse design and optimization process with specified performance criteria as inputs, along with the verification process. Specifically, the corresponding CP distributions are first generated through the diffusion model $D$ based on the predefined six pressure performance features $P_{target}$:

$$CP_{generated} = D(P_{target}; \theta_D), \qquad (2)$$

where $\theta_D$ is the parameter of the diffusion model. Subsequently, the generated CP distribution is transformed into the corresponding geometric airfoil parameters using the mapping model $M$.

$$A_{design} = M(CP_{generated}; \theta_M), \qquad (3)$$

where $\theta_M$ is the parameter of the mapping model. In this approach, the loss values of the diffusion model and the mapping model do not directly determine the accuracy of the generated airfoils, rather, the key lies in whether the actual performance of the generated airfoils meets the target indexes after validation by CFD. Therefore, the validity of the mapping model needs to be assessed by analyzing the difference between the post-simulation pressure distribution $CP_{CFD}$ and the model input pressure distribution $CP_{generated}$.

$$\Delta CP = \|CP_{CFD} - CP_{generated}\|. \qquad (4)$$

The overall effectiveness of the inverse design framework is then evaluated by measuring the difference between the target performance indicators $P_{generated}$ and the post-simulation performance indicators $P_{CFD}$.

$$\Delta P = \|P_{CFD} - P_{generated}\|. \qquad (5)$$

In addition, to improve design efficiency, the EGO optimization algorithm is introduced to directly optimize the performance indicators. An active learning strategy is integrated into the optimization process, whereby the optimized airfoil shape is reintroduced into the model framework for dynamic updating. This approach progressively reduces the design search space, enabling a more rapid identification of a feasible airfoil shape, as shown in Eq. (6). $\alpha(M(D(P)); \mathcal{D})$ denotes the optimization function under the current dataset $\mathcal{D}$ for selecting the next optimal airfoil $A_{opt}$.

$$A_{opt} = arg \min_{A} \alpha\left(M(D(P)); \mathcal{D}\right). \qquad (6)$$

This inverse design framework is designed to assist designers in the continuous optimization process to quickly and accurately generate airfoils that meet requirements, while reducing trial and error costs and time consumption in the design process. It is worth noting that this framework is not only suitable for the inverse design of airfoils, but also for the design verification of other parts or products.

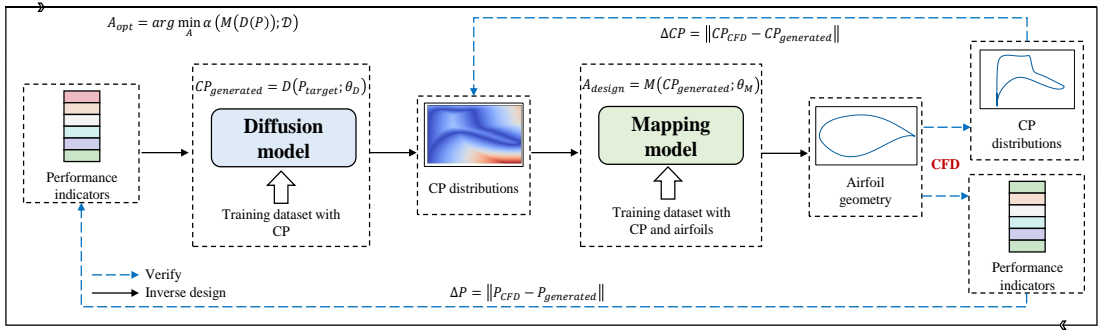

**Fig. 1** Airfoil inverse design method and verification.

## 3 Airfoil inverse design framework based on a classifier-free guided denoising diffusion probability model

This section introduces the airfoil inverse design framework based on a classifier-free guided denoising diffusion probabilistic model. First, the basic theory of the diffusion model is briefly discussed, with the conditional generation of the corresponding CP distribution based on pressure features. Subsequently, the mapping model converts the CP into the corresponding airfoil geometry. Finally, the overall framework is dynamically optimized by integrating EGO and active learning strategies.

### 3.1 Diffusion model

Motivated by the principles of non-equilibrium thermodynamics, the fundamental concept of the DDPM (Ho et al. 2020) involves generating data by gradually "diffusing" noise and subsequently training a neural network to reverse this diffusion process. The diffusion process resembles a Markov chain that incrementally introduces noise to the data in the opposite direction of sampling until the signal is destroyed. Within the DDPM framework, the addition of noise and the denoising process are defined as the forward and reverse processes, respectively.

In the forward process, Gaussian noise is gradually added to the given input data $x_0 \sim q(x)$ according to the time step $t$, as described in Eq. (7).

$$q(x_t|x_{t-1}) = \mathcal{N}(x_t; \sqrt{1-\beta_t}x_{t-1}, \beta_t I), \tag{7}$$

where $x_t$ is obtained by adding Gaussian noise to $x_{t-1}$ at time step $t \in \{T, T-1, \cdots, 1\}$. The predefined variance $\beta_{1:T}$ for $t$ is typically small and follows an increasing pattern. For instance, Jonathan Ho et al. (Ho et al. 2020) define $\beta_{1:T}$ as a linear function with values ranging from $10^{-4}$ to $0.02$. $\sqrt{1-\beta_t}$ serves as the scaling parameter, which diminishes as the time step increases. In order to minimize redundant iterations, a closed-form expression is derived using reparameterization techniques based on the characteristics of Gaussian distribution, and $x_t$ is directly calculated through $x_0$.

$$q(x_t|x_0) = \mathcal{N}\left(x_t; \sqrt{\bar{\alpha}_t}x_0, (1-\bar{\alpha}_t)I\right), \tag{8}$$

$$x_t = \sqrt{\bar{\alpha}_t}x_0 + \sqrt{1-\bar{\alpha}_t}\epsilon, \tag{9}$$

where $\alpha_t = 1 - \beta_t$, $\bar{\alpha}_t = \prod_{i=1}^{t}\alpha_i$ and $\epsilon \sim \mathcal{N}(0, I)$. When $T$ becomes large, $\bar{\alpha}_t$ converges towards zero, and the final distribution of $x_T$ gets nearer to the standard normal distribution. The process of forward diffusion stops when the final distribution becomes disordered enough for it to be viewed as an isotropic Gaussian distribution.

In the reverse process, $x_t$ is gradually denoised through $T$ time steps to obtain $x_0$. The reverse denoising process is defined as Eq. (10) and Eq. (11).

$$p_\theta(x_{t-1}|x_t) = \mathcal{N}\left(x_{t-1}; \mu_\theta(x_t, t), \sum\nolimits_\theta(x_t, t)\right), \tag{10}$$

$$x_{t-1} = \mu_\theta(x_t, t) + \sigma_t z, \tag{11}$$

where $\sigma_t$ represents the standard deviation, and $z \sim \mathcal{N}(0, I)$. In practice, the variance $\sum_\theta(x_t, t)$ is typically set as a constant, and the average value $\mu_\theta(x_t, t)$ is acquired through a neural network that is $\theta$-parameterized. The reverse process is derived using the idea of maximizing evidence lower bound (ELBO) of variational autoencoders, and the best form of mean $\mu_\theta(x_t, t)$ parameterization is obtained:

$$\mu_\theta(x_t, t) = \frac{1}{\sqrt{\alpha_t}}\left(x_t - \frac{\beta_t}{\sqrt{1-\bar{\alpha}_t}}\epsilon_\theta(x_t, t)\right). \tag{12}$$

The complete sampling process is similar to Langevin dynamics, where $\epsilon_\theta$ is the noise that needs to be predicted. The diffusion model in this article adopts a modeling method that predicts noise. The training goal is to reduce the gap between actual and predicted noise by optimizing the negative log-likelihood using a variational lower bound. The simplified loss function is as Eq. (13).

$$L(\theta) = E_{t\sim[1-T], x_0\sim q(x), \epsilon\sim \mathcal{N}(0,I)}[\|\epsilon - \epsilon_\theta(x_t, t)\|^2]. \tag{13}$$

Although DDPM has excellent performance in generating unconditional images, additional improvements are needed for generating conditional images. According to Dhariwal and Nichol (Dhariwal and Nichol 2021), under classifier guidance, sample data can be produced that fulfil the specified requirements. Specifically, the mean $\mu_\theta(x_t|y)$ and variance $\sum \theta(x_t|y)$ of the diffusion model is perturbed by the gradient of the classifier $p_\varnothing(y|x_t)$ to the target class $y$.

$$\hat{\mu}_\theta(x_t|y) = \mu_\theta(x_t|y) + s\sum \theta(x_t|y)\nabla_{x_t} log\left(p_\varnothing(y|x_t)\right). \tag{14}$$

The coefficient $s$ serves as a guidance index to assess the quality and diversity of control samples (greater s indicates superior quality and lower diversity). Although classifier guidance can generate targeted images, it introduces certain challenges because it requires training a separate classifier using noisy input images. Consequently, it is not feasible to use a standard pre-trained classifier, and the introduction of a classifier adds additional complexity and computational effort.

Ho and Salimans (Ho and Salimans 2022) first proposed a classifier-free guidance method that does not require a separate classifier. The predicted target category $y$ is intermittently used and randomly replaced with empty labels at a fixed probability. Therefore, the model can be used for unconditional generation and conditional generation, and its linear combination is Eq. (15).

$$\hat{\epsilon}_\theta(x_t|y) = (1+\omega)\epsilon_\theta(x_t|y) - \omega\epsilon_\theta(x_t), \tag{15}$$

where the implied-classifier weights $\omega$ is a guidance scale that used to generate the model along the $\epsilon_\theta(x_t|y)$ direction. By utilizing classifier-free guidance, a single model can leverage its own knowledge without the requirement for separate classifications.

The classifier-free guided diffusion model is adopted as the generative model. Compared to generative models such as GANs and VAEs, the diffusion model is constructed based on physical and mathematical principles, enabling more accurate identification of aerodynamic features. Its training process is more stable and reliable, less susceptible to issues such as gradient vanishing and model collapse. The network framework is shown in Fig. 2, which is mainly divided into two stages: training and sampling.

During the training phase of the diffusion model, processed CP distributions and pressure features are employed as training data. Specifically, the six pressure feature indicators are utilized as diffusion conditions, and the UNet network is employed to predict noise during the diffusion process ϵ. The initial CP distribution $x_0$ undergoes a process of gradual Gaussian noise addition, forming multiple intermediate states $x_t$. In this process, pressure features are transformed and conditioned through a linear network, then used as the input of the UNet network together with the time step $t$ and the corresponding intermediate state $x_t$. The predicted noise $\epsilon_\theta$ outputted by the network and the actual added noise ϵ, are used to compute the loss function via Eq. (13), guiding the optimization updates of network parameters. In the sampling stage of the diffusion model, the implicit classifier-free guidance method is employed, utilizing Eq. (15) to jointly train conditional and unconditional models for step-by-step noise sampling. The Impact of implicit classifier weights $\omega$ is adjusted to control the proportion of input conditions and achieve a balance between precision and diversity. By gradually sampling the noise, CP distributions that conform to the specified pressure features are ultimately generated. The UNet network for noise prediction consists of three downsampling and three upsampling layers, where pressure feature conditions are transformed into corresponding dimensions and added to the model in an additive manner.

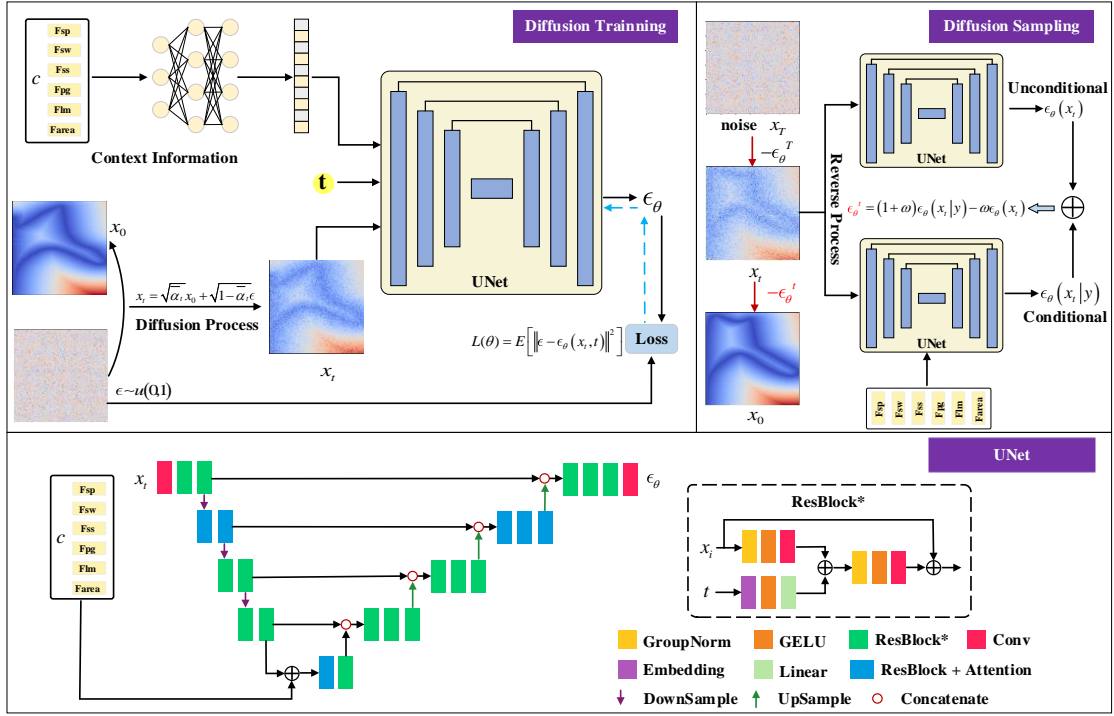

**Fig. 2** CDDPM network framework based on UNet prediction noise.

### 3.2 Mapping model

The mapping model primarily nonlinearly maps the CP to the geometry of the airfoil through a CNN, whose architecture is symbolically depicted in Fig. 3. According to prior research findings, employing a network structure consisting of three convolutional layers followed by a fully connected layer has been shown to yield satisfactory results (Deng and Yi 2023). Each convolutional layer mainly consists of convolution operation, batch normalization and activation function ReLU. Within the convolution layer, the convolution operation processes the CP within a 2D matrix by capitalizing on sparse interaction and parameter sharing principles. In this work, the model input is a $128 \times 128$ CP distribution. The output of the model consists of 130 points on the $y$ coordinate of the airfoil geometry, distributed with 65 points on the upper surface and 65 on the lower surface, while the $x$ coordinate remains constant.

$$x_i = \frac{1}{2}\left(\cos\frac{2\pi(i-1)}{130} + 1\right), i = 1,2,\cdots,131. \tag{16}$$

$$MSE = \frac{1}{m}\sum_{i=1}^{m}(\|y' - y\|^2). \tag{17}$$

The aim of training a mapping model is to update the parameter weights in order to minimize the loss function. In this instance, the optimization objective is to decrease the sum of the average error terms. The formula is as Eq. (17), where $m$ represents the number of samples.

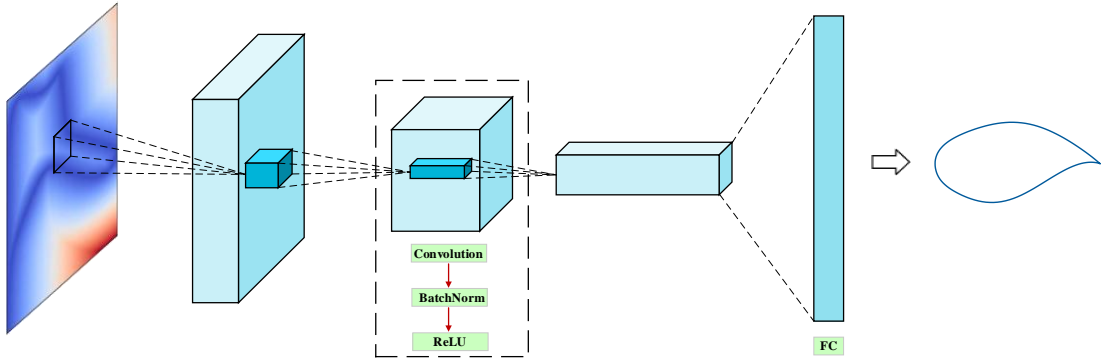

**Fig. 3** Architecture of mapping model.

**3.3 Optimization model**

Generally, it is challenging for designers to directly set performance indicators that are consistent with the physical flow laws and to ensure that the combination of performance indicators is self-consistent. Furthermore, even if reasonable performance indicators are established based on extensive historical experience, it is challenging to generate optimal airfoil geometries at once. Since both diffusion and mapping models are trained based on initial data, it is unlikely that optimization based on them will achieve the required accuracy immediately. Optimization studies based on generative models usually iteratively add the training data to improve the accuracy of the generated models, a technique known as an active learning strategy(Yang et al. 2023c). Therefore, a practical method to readjust each performance indicator value is proposed, which uses EGO to optimize the performance indicators and construct the optimization model in combination with the active learning strategy, thereby achieving the dynamic update of the inverse design framework.

The process framework of the optimization model is shown in Fig. 4. First, the diffusion model and mapping model are trained using data from the initial training dataset. After the initial training is completed, the performance indicators are optimized using EGO in combination with the training framework. In each optimization iteration, new airfoils are generated. When the error criteria for the performance indicators of the generated airfoils are not satisfied, these data are incorporated into the training dataset, and training of the diffusion and mapping models continues. This active learning strategy will continue until the error between the performance indicators of the airfoils generated by the inverse design framework and those of the numerical solver is reduced to meet the predefined error criteria. Finally, the EGO-optimized performance indicators are used as inputs to the framework to generate the optimal airfoil geometry.

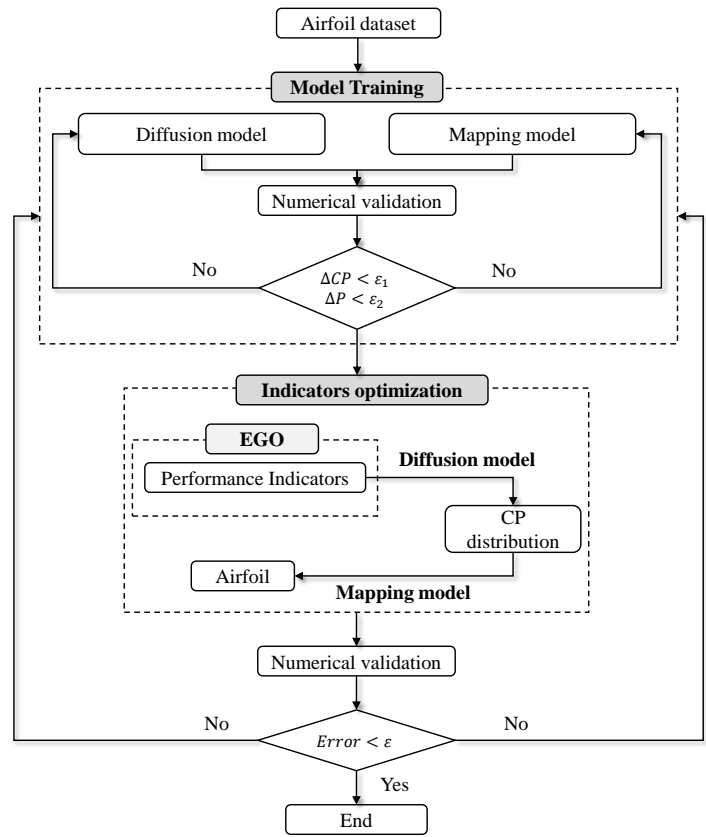

**Fig. 4** Process framework for the optimization model

## 4 Experimental design and performance analysis

This section first describes the framework of the process for constructing the airfoil dataset and uses scripts to automate its construction. Subsequently, the performance analysis of the diffusion and mapping models in the above methods is carried out to assess whether their accuracy is sufficient for practical design applications. The overall network framework is implemented using PyTorch (2019), and all parameters are optimized using the gradient-based optimizer Adam (Kingma and Ba 2017). The experiments are conducted on the NVIDIA GeForce RTX 3090 GPU platform.

### 4.1 Dataset preparation

The model requires training and validation based on data to ensure the quality of generation. To guarantee the reliability and accuracy of the dataset, Fig. 5 shows the procedural schematic for constructing a dataset that includes airfoil geometric shapes, CP distributions, and their corresponding features.

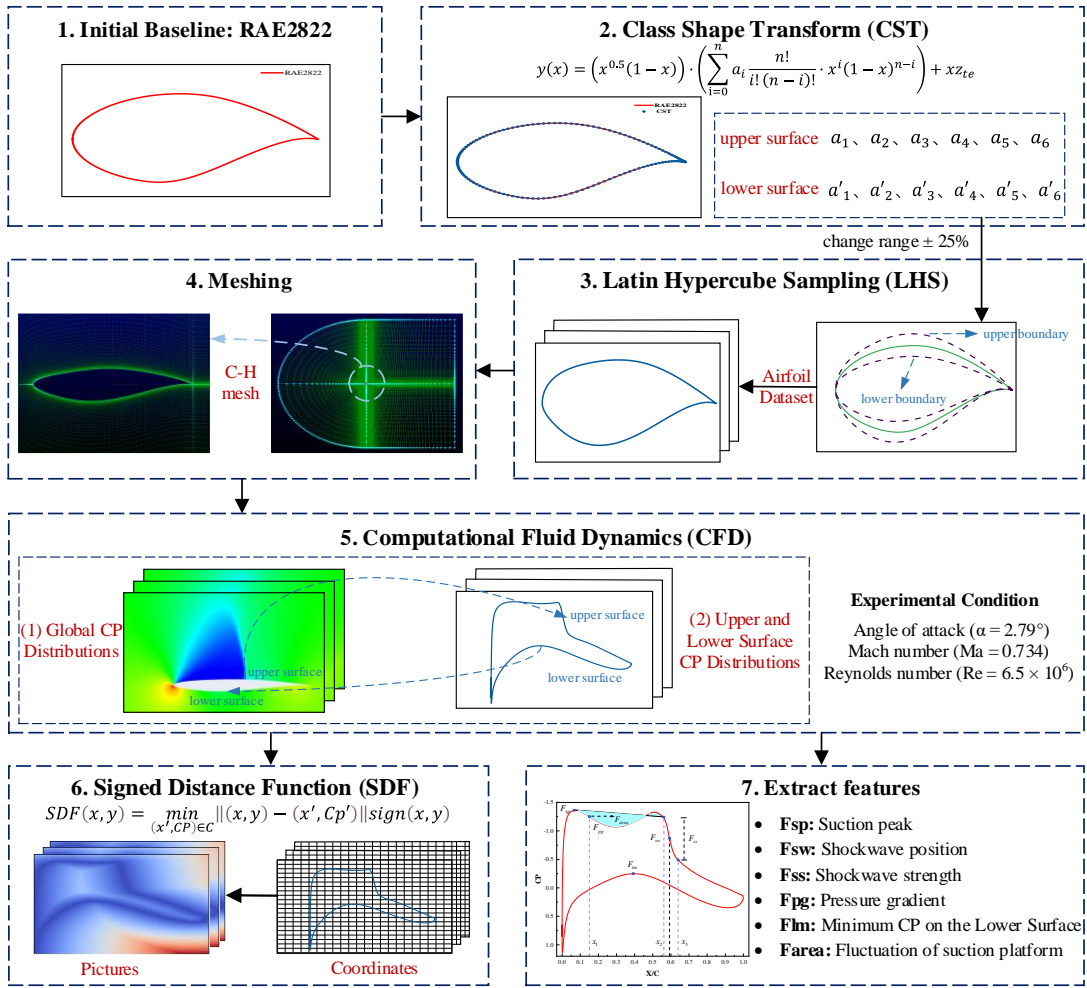

**Fig. 5** Schematic of airfoil dataset building, including the transformation of CP distributions and features extraction.

The dataset uses the RAE2822 supercritical airfoil as the initial baseline and applies the 6th order Class Shape Transformation (CST) method(Wu et al. 2019), described by Eq. (18), to fit and derive the baseline airfoil. Herein, $x$ and $y$ represent the transverse and longitudinal coordinates of the airfoil, respectively; $a_i$ denotes the parameters to be varied subsequently, $n$ indicates the order of the CST function, and $z_{te}$ denotes the trailing-edge half-thickness. It is worth noting that the variation in CST parameters is limited to within 25%, which is considered a reasonable range. The Latin Hypercube Sampling (LHS) method is utilized to generate a set of N airfoils within the designated design space, where N is set to 1000. Each airfoil consists of 130 points represented as (x, y) coordinate tuples, arranged in a counterclockwise direction starting from the trailing edge. It should be mentioned that the x-coordinates remain consistent across all airfoils.

$$y(x) = \left(x^{0.5}(1-x)\right) \cdot \left(\sum_{i=0}^{n} a_i \frac{n!}{i!\,(n-i)!} \cdot x^i (1-x)^{n-i}\right) + xz_{te}. \tag{18}$$

$$y = CST\bigl(LHS(a_i, N)\bigr). \tag{19}$$

Subsequently, the airfoil underwent a mesh division using the C-H type mesh approach, resulting in a partition into 73,904 individual meshes. The aerodynamic analysis of the airfoil dataset was conducted under prescribed conditions, including a fixed Reynolds number (Re = $6.5 \times 10^6$), Mach number (Ma = 0.734), and angle of attack ($\alpha = 2.79°$). These conditions align with those outlined in Case 9 of Cook et al.'s work (Cook 1977). The flow fields are obtained by solving the Reynolds-averaged Navier-Stokes (RANS) equations using the finite volume method. As illustrated in Fig. 6, the results of the CFD simulations agree well with the wind tunnel experimentation, thereby meeting the requirements for experimental validation.

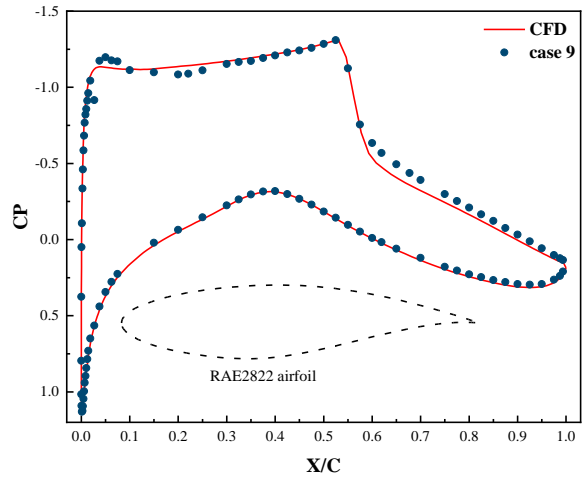

**Fig. 6** Comparison of simulated pressure coefficient distributions with implementation results.

In this work, due to the discrete nature and relatively limited quantity of pressure coordinates, they are unsuitable for model generation and features extraction. Consequently, before training, the CP distribution samples are transformed into pixel image data represented as a 2D matrix. The selected geometric representation utilizes the Signed Distance Function (SDF) sampled on a Cartesian grid (Wang et al. 2023). The use of SDF offers a versatile representation for CP distributions exhibiting diverse characteristics and facilitates effective alignment with diffusion models. The SDF for each pixel within the matrix is defined as the minimum distance from the pressure curve $C$, and its mathematical formulation can be expressed as Eq. (20).

$$SDF(x,y) = \min_{(x',CP)\in C} \|(x,y) - (x', Cp')\| sign(x,y). \tag{20}$$

The pressure distribution is represented by a $128 \times 128$ Cartesian grid within $(x, y) \in [0,1] \times [-1.5, 1.15]$. $sign(x, y)$ determines whether the pixel is inside or outside the curve. If $(x, y)$ is located in the area enclosed by the curve, then $sign(x, y) = 1$, otherwise $sign(x, y) = -1$. Here $sign(x, y) = 1$ to better observe the geometry of the pressure.

In the process of designing airfoils, it is essential to account for a range of design objectives and constraints, including factors such as lift, drag, and minimizing the stall angle. However, comprehensively accounting for all design elements can lead to a significant increase in design complexity and resource allocation. Research indicates that the pressure distribution across the airfoil surface encapsulates valuable information pertaining to its aerodynamic characteristics. Consequently, the performance-oriented design of the airfoil can be indirectly guided by imposing constraints on key pressure distribution characteristics, thereby improving the overall efficiency of the design process. Building upon previous works (Li et al. 2018), this paper uses six distinctive pressure coefficient features widely recognized in the field of supercritical airfoil aerodynamic design, as shown in step 7 in Fig. 5 and described as follows:

Suction peak $F_{sp}$ : This is value located at the point with minimum pressure coefficient near the leading edge of the airfoil. The peak pressure suction should be maintained within a certain range, avoiding excessively high values while ensuring it does not fall below -1.0. This constraint aims to mitigate excessive shockwave strength, maintain a reasonable leading-edge radius, and prevent the airfoil from adopting a "peak" configuration. Specifically, the suction peak is determined by evaluating the minimum pressure coefficient value within the initial 15% of the spanwise distance along the upper surface, where $x_1 = 0.15$.

$$F_{sp} = \min_{x \leq x_1} CP_{up}(x). \tag{21}$$

Shockwave position $F_{sw}$ : This parameter represents the maximum rise ratio of CP on the upper surface of the airfoil. Typically, the position of the shockwave is located near the rear of the center chord, which is beneficial for reducing drag and maintaining the stability and high-speed performance of the aircraft. The abscissa of the shockwave's end is denoted as $x_3$.

$$F_{sw} = \max_{x_1 \leq x \leq x_3} dCP_{up}/dx. \tag{22}$$

Shockwave strength $F_{ss}$ : Although previous studies have employed various definitions for shockwave strength, it can be broadly defined as the pressure increase between the two sides of the shockwave. This metric directly characterizes the airfoil's drag performance. The horizontal coordinate of the shockwave's starting position is specified as $x_2$.

$$F_{ss} = CP_{up}(x_3) - CP_{up}(x_2). \tag{23}$$

Pressure gradient $F_{pg}$ : Ideally, the pressure gradient, extending from the suction peak to the front of the shockwave, should be within the range of -0.2 to 0.5. This range ensures that the suction platform has the necessary length to provide adequate lift. It is important to note that an excessively high-pressure gradient can cause the pressure to surpass the fluid's inertial forces, leading to the fluid deviating from the airfoil's contour.

$$F_{pg} = (CP_{up}(x_2) - CP_{up}(x_1))/(x_2 - x_1). \tag{24}$$

Minimum CP on the Lower Surface $F_{lm}$ : This parameter denotes the lowest value of the pressure coefficient observed on the lower surface of the airfoil (Botero-Bolívar et al. 2023). To prevent the occurrence of supersonic conditions on the lower surface, it is essential for the negative pressure to exceed -0.35.

$$F_{lm} = \min_{0 \leq x \leq 1} CP_{low}(x). \tag{25}$$

Fluctuation of suction platform $F_{area}$ : This parameter is defined as the area enclosed by the blue zone located between the suction peak and the shockwave on the airfoil's upper surface. The fluctuation of the suction platform serves as a metric for quantifying the extent of pressure platform tortuosity. The pressure platform should be as smooth as possible to minimize undesirable robustness issues. Where $f_l(x)$ represents the linkage between the suction peak and the shockwave.

$$F_{area} = \int_{x_1}^{x_2} \left| f_{CP_{up}}(x) - f_l(x) \right| dx. \tag{26}$$

**4.2 Performance analysis**

4.2.1 The impact of airfoil numerical values on mapping model

During the training process of the mapping model, the network trains using 90% of the samples, while the remaining 10% are used to verify the model's performance. The hyperparameters are based on past research (Hui et al. 2020), with the batch size set to 64, training epochs of 300, and convolution kernel size of 5. At the start of training, the learning rate is set at $1.0 \times 10^{-4}$ and gradually decreases as the number of training rounds increases, promoting quicker network convergence. The training process of CNN involves optimizing the network through backpropagation, which updates its parameters. Therefore, Eq. (17) is used as the loss function to evaluate the convergence of the mapping model.

After 300 training epochs, the loss value of the mapping model converges to $4.8 \times 10^{-7}$, making it difficult to further enhance accuracy through adjustments to the model structure. Although the current accuracy surpasses that of previous researches(Lei et al. 2021; Deng and Yi 2023), there remains a certain degree of deviation in the simulated airfoil after simulation. Subsequent experiments revealed that smaller numerical values may lead to potential data loss during the training process. The value of the airfoil's y-coordinate falls within a range of $\pm 0.08$, where the y-coordinate near the leading and trailing edges is close to zero. Tiny values could result in data loss during model training. Therefore, the y-coordinate is expanded by 1000 times based on the original airfoil to mitigate the impact of accuracy issues caused by excessively small values. Figure 7 shows the impact of expanding the y-coordinate values on the convergence of the mapping model. To ensure fair comparison, loss values of the same magnitude are used. After enlarging the airfoil numerical values, although the training period increased, the loss value significantly decreased, leading to a more pronounced improvement in results.

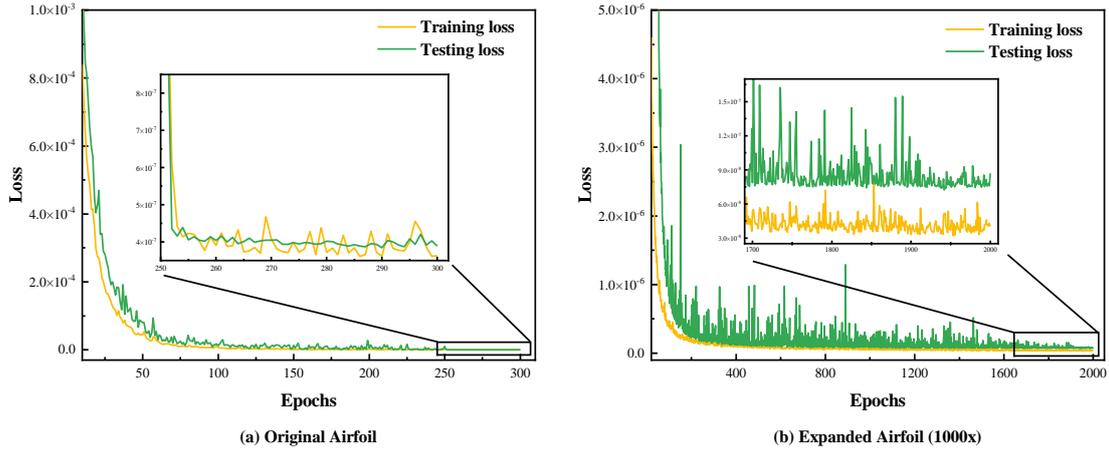

Fig. 7 Convergence history of the loss function between original airfoil and expanded airfoil.

To further investigate the model's mapping results, the mean absolute error (MAE) and mean relative error (MRE) were utilized to assess the accuracy of the predictions for each point along the x-axis of the airfoil in the test set.

$$MAE_y = \frac{1}{m}\sum_{i=1}^{m}\left|\hat{y}^i{}_j - y^i{}_j\right|, \qquad (27)$$

$$MRE = \frac{1}{m}\sum_{i=1}^{m}\frac{\left|\hat{y}^i{}_j - y^i{}_j\right|}{max(\left|y^i{}_j\right|,\left|\hat{y}^i{}_j\right|)}. \qquad (28)$$

As shown in Fig. 8, both absolute errors fall within the tolerance requirements proposed by Sobieczky (Sobieczky 1999). Specifically, the fitting error of the airfoil parameterization is less than 0.0007, meeting the tolerance zone of the wind tunnel experiment. Consequently, the designed mapping model can accurately reflect the mapping relationship between the pressure map and the airfoil, and capture the nonlinear characteristics in fluid mechanics. Furthermore, enlarging the numerical values of the airfoil resulted in a reduction of the absolute error values by almost half, particularly with significant improvements in the lower surface and leading-edge regions. MRE for each point on the airfoil is depicted in Fig. 9. Compared to the original airfoil, the prediction accuracy of the expanded airfoil numerical values has greatly improved. The MRE of the first half of the airfoil fluctuates around 0.005. It is worth noting that there is some fluctuation in the position where the value changes from negative to positive on the underside of the airfoil, causing a sudden increase in the MRE at multiple points located at 0.9 on the x-axis. However, the results of subsequent verification show that linear fitting of the curve can mitigate the impact of this occurrence on results. Overall, expanding the airfoil numerical values can considerably improve the accuracy of the results. The high-precision prediction results also ensure that the mapping model can accurately and reliably complete the mapping from the pressure distributions to the airfoil geometry.

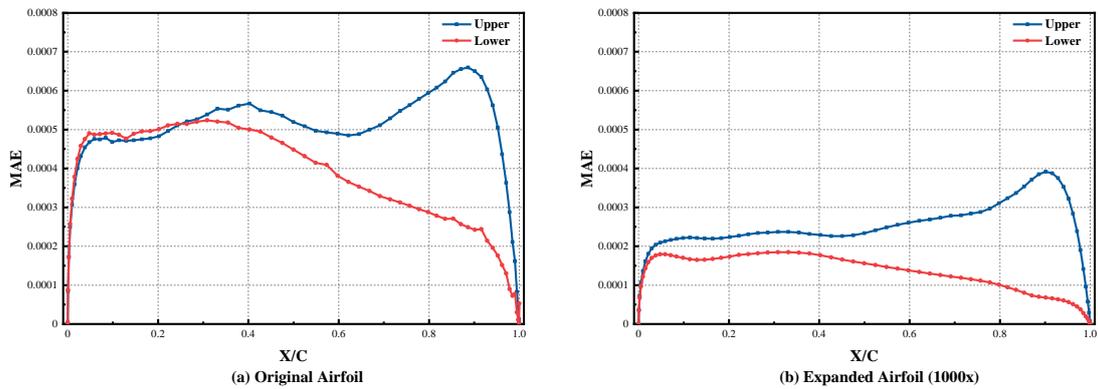

**Fig. 8** Absolute error between original airfoil and expanded airfoil.

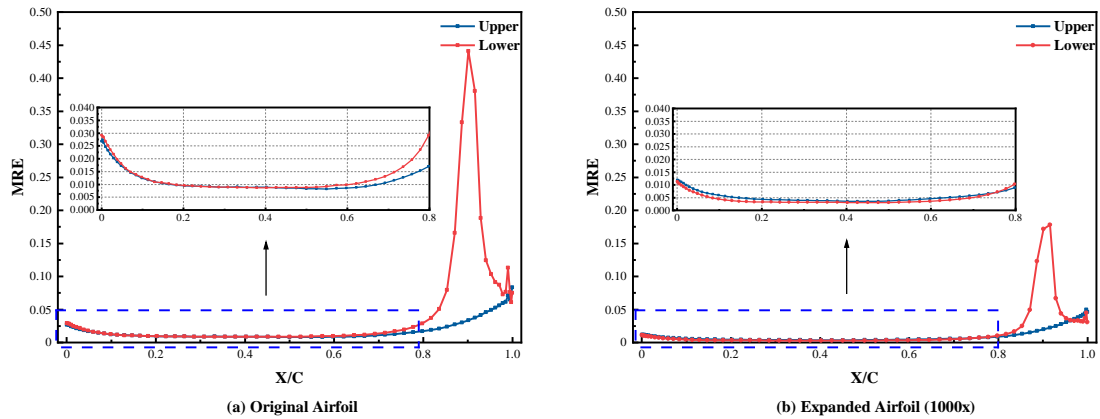

**Fig. 9** MRE between original airfoil and expanded airfoil.

Pressure distributions are randomly selected from the test set to confirm the effectiveness of the mapping model. As illustrated in Fig. 10, the mapping model successfully recognizes the corresponding airfoil geometry from the pressure map, generating prediction results that are nearly indistinguishable from the actual airfoil. To analyze the flow field of the mapped airfoil using the same network topology and CFD solver, spline curve fitting is performed on the mapping results. This process eliminates the problems of local incomplete smoothness and trailing edge coordinate intersections. The verification results indicate that the pressure distribution of the mapped airfoil closely aligns with the original distribution. A high-precision mapping model will significantly enhance the practical effectiveness of inverse design.

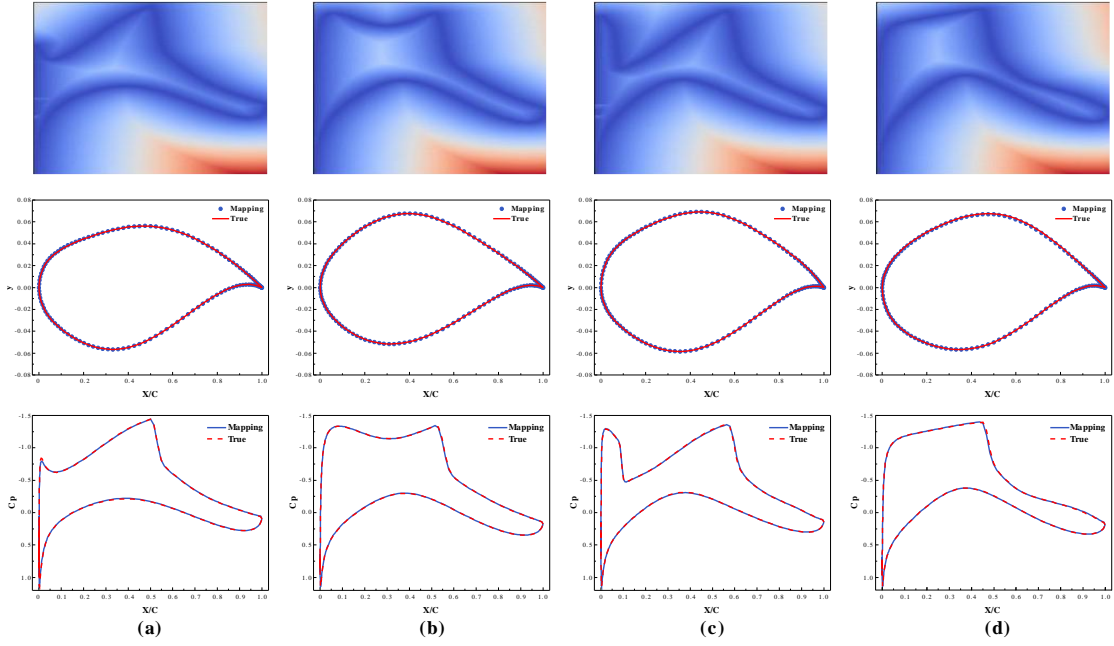

(a) (b) (c) (d)

**Fig. 10** Typical predicted result by mapping model from different CP distributions.

4.2.2 The impact of implicit classifier weights on generated results

The diffusion model adopts a classifier-free guidance method to balance the impact of conditions on the generated results, similar to classifier guidance or GAN truncation. To optimize the CDDPM for accurately generating corresponding pressure distributions based on specified CP features, a detailed analysis of the implicit classifier weights was $\omega = \{0.1, 0.2, 0.3, \cdots, 4, 5\}$ conducted. During training, the step size $T$ is set to 400, the batch size to 12, and $\beta_1 = 10^{-4}$ linearly increases to $\beta_T = 0.02$ in accordance with the step size. The initial learning rate is set at 1e-4, using the Adam optimizer to train the MSE loss function. As the training epochs increases, the learning rate is gradually decreased to accelerate the convergence of the model. Currently, it takes about 28 hours to train the model for 2,500 epochs, and only 20 seconds to generate a sample of a specific airfoil. For training our model, 95% of airfoil data was used, and 10 new airfoil data were randomly generated using the remaining untrained pressure feature data to determine the optimal $\omega$ value.

In this work, pressure distributions generated from specified pressure features are mapped to the geometric shapes of airfoils, and high-fidelity CFD simulations are then utilized to validate the results of the inverse design. The impact of implicit classifier weights $\omega$ on the generated results in the airfoil test set is shown in Fig. 11. To accurately measure the results, the performance indicators of mean absolute error is used for evaluation.

$$MAE_f = \frac{1}{m}\sum_{i=1}^{m}|F_{feature} - F_{feature}'|. \tag{29}$$

$F_{feature}$ and $F_{feature}'$ represent the specified CP features and the CP features obtained through verification respectively. Experimental results indicate that the error is minimized when $\omega = 1.0$. This suggests that $\omega$ can more reasonably weigh the proportions of the unconditional and conditional distributions at this time, thereby generating the corresponding airfoil shapes.

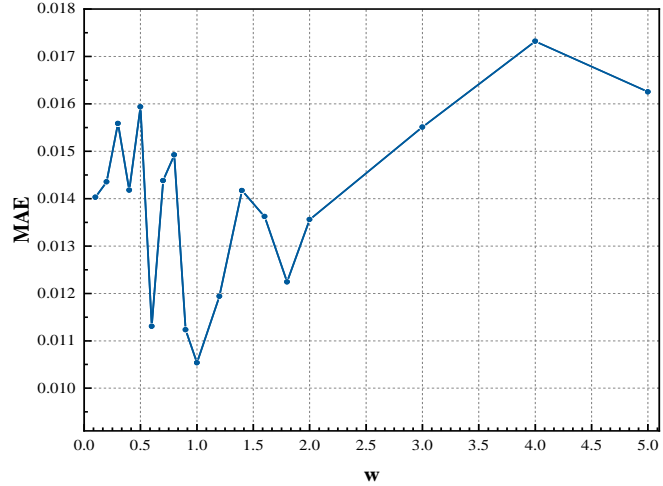

**Fig. 11** Impact of implicit classifier weights on generated results.

Fig. 12 shows the pressure distributions generated by our model under random generation conditions and when $\omega = 1.0$. This figure distinctly illustrates the significant impact of $\omega$ on the generation process of the diffusion model. Increasing the implicit classifier weight has the effect of reducing sample diversity and improving the fidelity of a single sample. For airfoil inverse design, precise conditional generation is far more critical than sample diversity. The model demonstrates its advantage in practical applications only when it can accurately generate airfoils that match specific CP features.

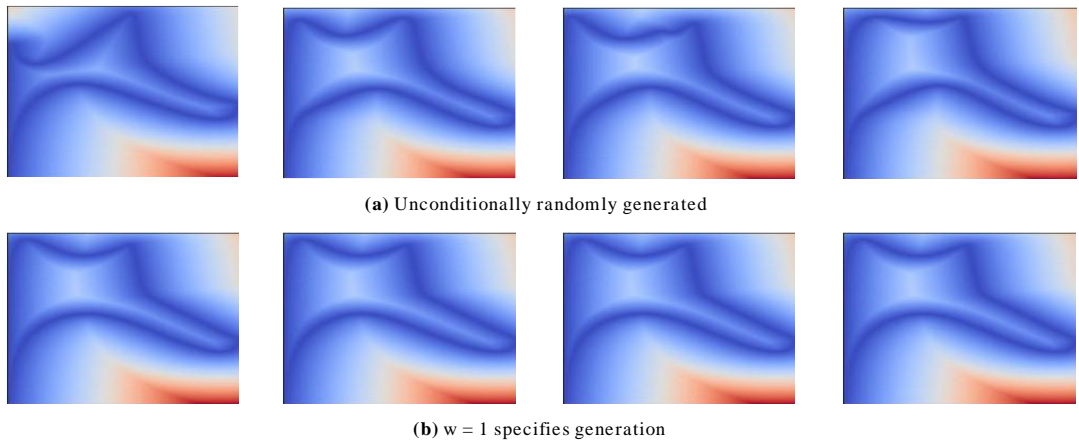

**Fig. 12** Repeat four times to generate a pressure distribution under a given CP features.

## 5 Results and discussions

This section evaluates the effectiveness of the method proposed in this paper by comparing it with existing methods for inverse designing airfoils. By thoroughly analyzing the coupling relationship between the performance indicators, the method aims to guide designers in the airfoil design process and validate the feasibility of the method in practical applications. Finally, based on the optimization approach to find the optimal airfoil shape under the constraints of pressure features, the trade-off considerations of designers on the combination of performance indicators are reduced.

### 5.1 Comparison verification

In order to further evaluate the effectiveness of the airfoil inverse design method proposed in this paper, we conducted a comprehensive comparison of currently existing inverse design methods. All of them adopt the inverse design framework outlined in Section 2, utilizing six pressure features as outputs to generate corresponding airfoils. We analyze the actual results based on CVAE, CGAN, WGAN, and the CDDPM proposed in this paper using the test set, respectively. Fig. 13 demonstrates the distribution of CP versus the actual CP obtained from the airfoils generated by each inverse design method based on the six pressure features and verified by CFD. Among them, the black solid line indicates the CP distribution for the specified pressure features; the dashed line with circular scatters indicates the design result of CVAE; the dashed line with triangular scatters indicates the design result of CGAN; the dashed line with square scatters indicates the design result of WGAN; and the dashed line with diamond scatters indicates the design result of CDDPM. As can be seen from Fig. 13, the distributional differences are more pronounced on the lower surface of the airfoil, which may be caused by the fact that there is only one $F_{lm}$ pressure constraint on the lower surface, but the $F_{lm}$ as a whole is still constrained to the minimum CP value on the lower surface. Compared with other methods, the CDDPM-based inverse design results are closer to the specified pressure features, and the validated CP distributions largely coincide with the actual distributions.

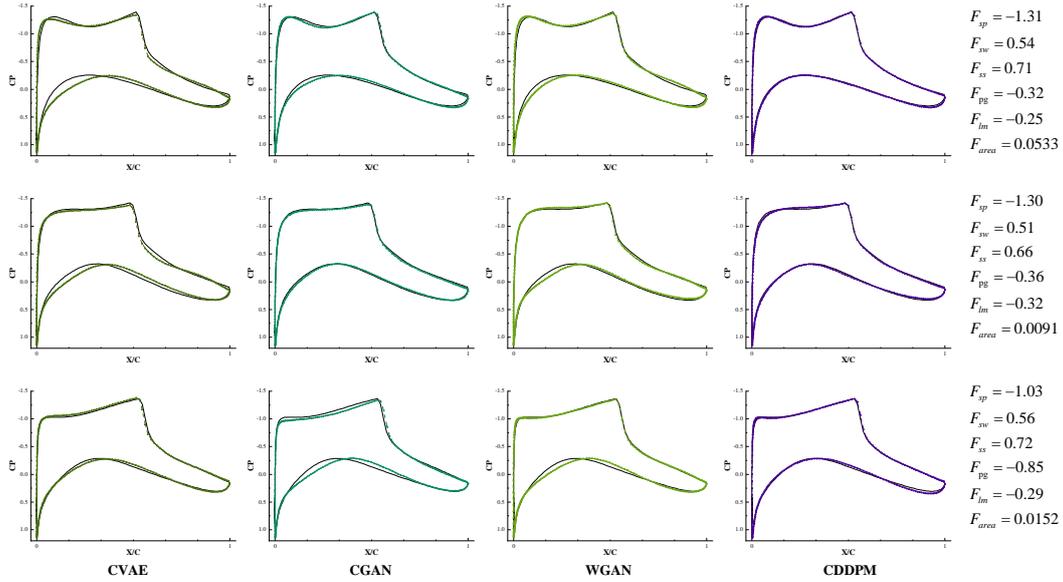

**Fig. 13** Verification of the design results of different inverse design methods according to the specified CP features

In addition, two metrics, absolute error and average relative error, were chosen to calculate the performance of each method on the test set. Table 1 shows the errors between the obtained pressure features and the specified pressure features of the airfoils generated by each inverse design method after CFD validation. Compared to $F_{pg}$ and $F_{area}$, these models perform better on $F_{sp}$, $F_{sw}$, $F_{ss}$, and $F_{lm}$, which is consistent with the results in Fig. 13. This is mainly due to the fact that $F_{pg}$ and $F_{area}$ themselves have a larger range of fluctuations and numerical differences, thereby making it more challenging for the models to capture the relevant features. Compared to the current more advanced generation method WGAN, CDDPM is only slightly less accurate in terms of feature $F_{ss}$, but overall, its accuracy is improved by 33.6%. CGAN and WGAN were fine-tuned several times during the training process but still faced problems of overtraining and model crashes, especially for CGAN. Furthermore, as shown in Fig. 14, we focus on analyzing the MAE distribution of the CDDPM-based inverse design method for each pressure feature on the test set. Under the specified pressure features, the models fit well with these features on most of the test samples without significant fluctuations, and the errors are all kept within 0.05. Overall, as shown in Table 1, our model has higher accuracy in identifying different pressure features compared to other generative models, which can better meet the accuracy requirements of complex designs and achieve better inverse design results.

**Table 1**

Performance comparison of different inverse design methods on six CP features.

| | Fsp | | Fsw | | Fss | | Fpg | | Flm | | Farea | |
|---|---|---|---|---|---|---|---|---|---|---|---|---|
| | MAE | MRE | MAE | MRE | MAE | MRE | MAE | MRE | MAE | MRE | MAE | MRE |
| CVAE | 0.0273 | 0.0217 | 0.0077 | 0.0140 | 0.0295 | 0.0408 | 0.1034 | 0.2508 | 0.0192 | 0.0670 | 0.0164 | 0.4258 |
| CGAN | 0.0313 | 0.0267 | 0.0062 | 0.0112 | 0.0279 | 0.0388 | 0.0942 | 0.2518 | 0.0125 | 0.0376 | 0.0098 | 0.3241 |
| WGAN | 0.0168 | 0.0140 | 0.0062 | 0.0113 | **0.0251** | **0.0349** | 0.0863 | 0.2259 | 0.0089 | 0.0280 | 0.0088 | 0.2859 |
| CDDPM | **0.0115** | **0.0095** | **0.0053** | **0.0097** | 0.0259 | 0.0362 | **0.0493** | **0.1693** | **0.0033** | **0.0119** | **0.0055** | **0.2280** |

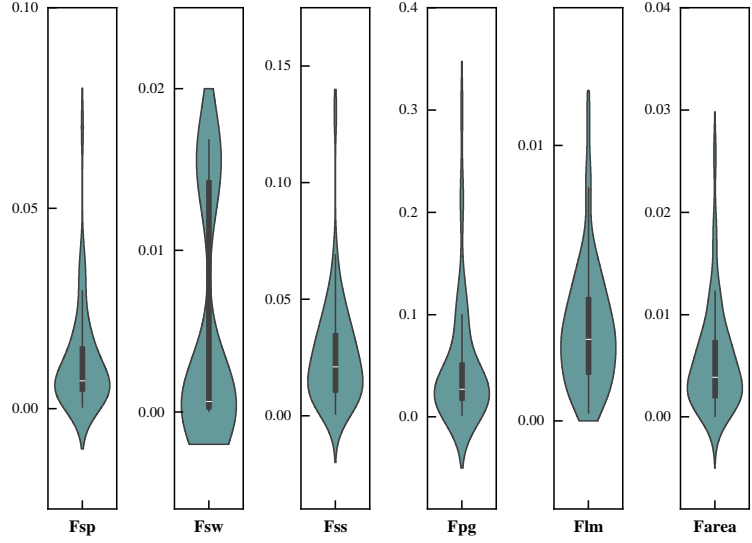

**Fig. 14** Distribution of MAE between different pressure features

To visually demonstrate the generative capability of CDDPM, Fig. 15 shows typical inverse design results of CDDPM on the test set, based on six specified CP features. CDDPM initially generates pressure distributions using SDF, which are then converted into corresponding geometric airfoils through the mapping model, followed by CFD solver validation of the fitted airfoils' CP features. The results show that the generated airfoils possess smooth and continuous characteristics, without significant fluctuations, and display distinct differences for varying CP features. The CFD fitting results highly align with the specified features. Although there is a small error, the accuracy of this method is sufficient for the preliminary design of the airfoils.

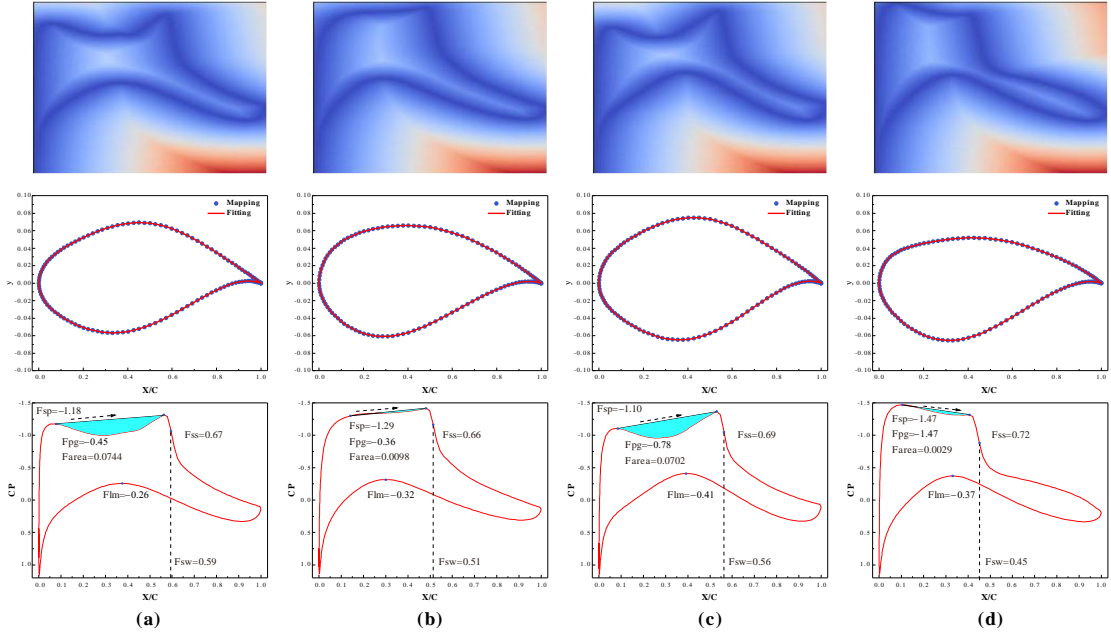

Fig. 15 Typical inverse design result by CDDPM from different CP. Specified CP features: (a) $F_{sp} = -1.17$, $F_{sw} = 0.59$, $F_{ss} = 0.68$, $F_{pg} = -0.45$, $F_{lm} = -0.26$, $F_{area} = 0.0765$,; (b) $F_{sp} = -1.30$, $F_{sw} = 0.51$, $F_{ss} = 0.66$, $F_{pg} = -0.36$, $F_{lm} = -0.32$, $F_{area} = 0.0091$; (c) $F_{sp} = -1.11$, $F_{sw} = 0.56$, $F_{ss} = 0.70$, $F_{pg} = -0.77$, $F_{lm} = -0.41$, $F_{area} = 0.0779$; (b) $F_{sp} = -1.47$, $F_{sw} = 0.45$, $F_{ss} = 0.72$, $F_{pg} = 0.57$, $F_{lm} = -0.37$, $F_{area} = 0.0029$.

**5.2 Coupling analysis of performance indicators**

In the actual design process of airfoils, designers usually do not start from scratch but rather improve upon existing designs. The relationship among different CP features in airfoils is highly nonlinear, and altering one feature often affects others. However, in the early stages of airfoil design, adjusting a specific feature allows for the quick observation of related derived changes. This is immensely beneficial for designers to identify coupling relationships among different CP features and identify directions for improvement.

Fig. 16 shows the changes resulting from altering different pressure features. These SDF pressure distributions are generated by changing only one of the features while the other five pressure features remain unchanged. Subsequently, these distributions were converted into corresponding geometric airfoils through the mapping model, followed by simulation and verification. The red line in the figure represents the CP distribution after CFD verification, closely matching the model's output. Fig. 16 (a) clearly shows that the model accurately identifies and reflects the decrease in $F_{sp}$ value on the generated pressure distributions. This confirms that the inverse design approach can achieve the same aerodynamic performance as the feature input, and the pressure distributions generated by the model are physically feasible.

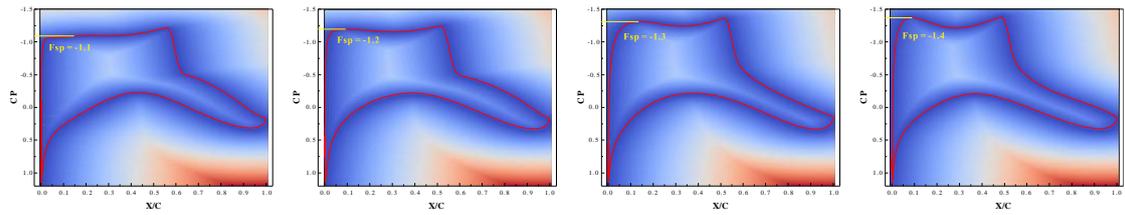
(a) The changing trend of CP after changing only Fsp. Fsp = -1.1, -1.2, -1.3, -1.4.

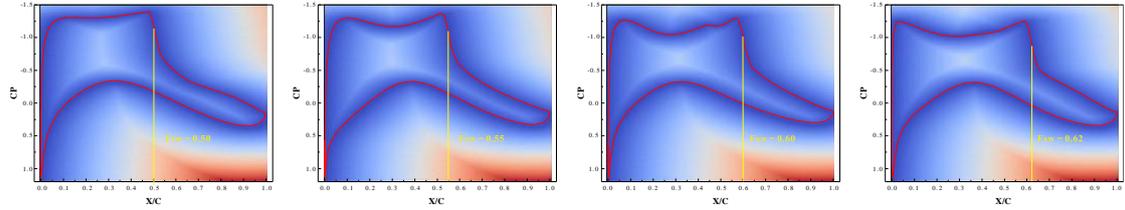
(b) The changing trend of CP after changing only Fsw. Fsw = 0.50, 0.55, 0.60, 0.62.

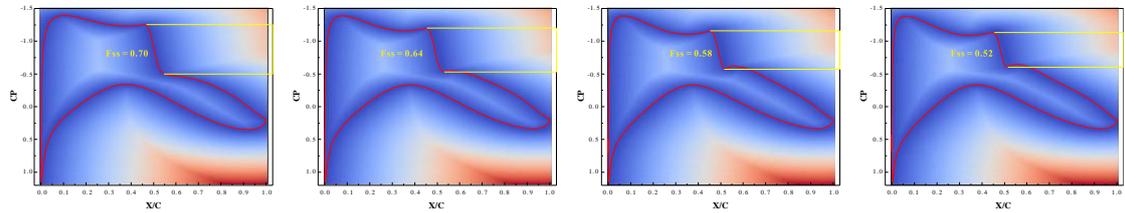
(c) The changing trend of CP after changing only Fss. Fss = 0.70, 0.64, 0.58, 0.52.

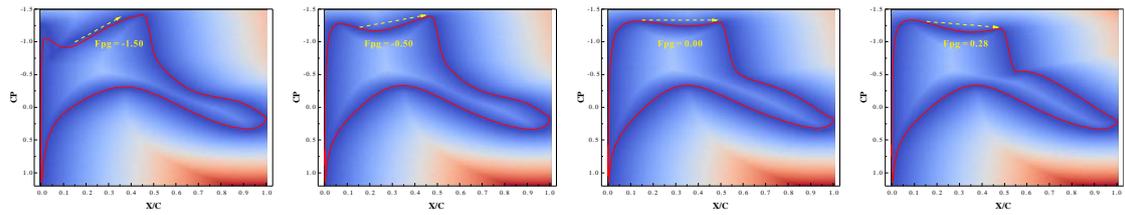
(d) The changing trend of CP after changing only Fpg. Fpg = -1.50, -0.50, 0.00, 0.28.

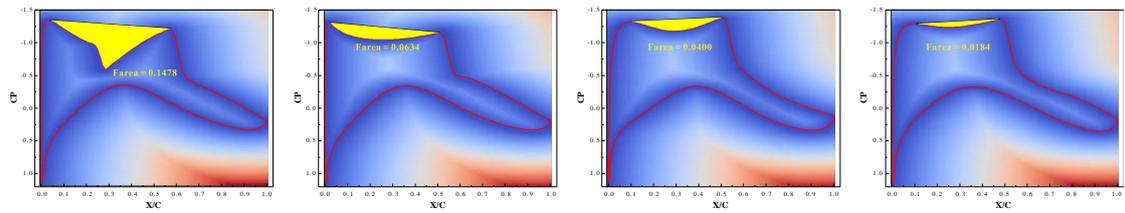
(e) The changing trend of CP after changing only Farea. Farea = 0.1478, 0.0634, 0.0400, 0.0184.

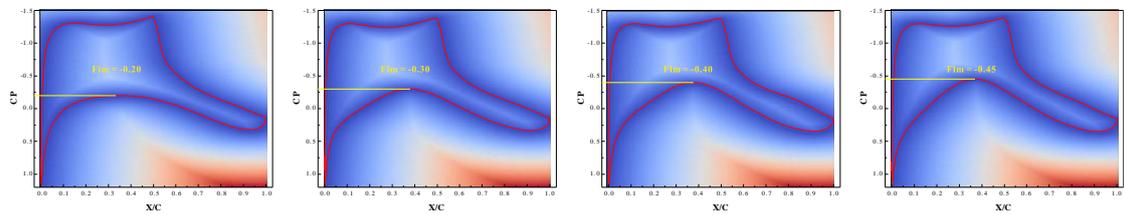
(f) The changing trend of CP after changing only Flm. Flm = -0.20, -0.30, -0.40, -0.45.

**Fig. 16** Airfoil inverse design results with gradual changes in pressure features.

In addition, increasing the height of the suction peak will lead to a corresponding increase in the pressure gradient. In addition, increasing the height of the inhalation peak will lead to a corresponding increase in the pressure gradient. Since the pressure gradient remained unchanged at the time of designation, this, in turn, will drive the shockwave position closer to the leading edge to minimize the overall change in pressure features resulting from this increase. Interestingly, changing the shockwave position has a minimal impact on the suction peak. Fig. 16 (b) shows two sets of inverse design results of gradually increasing $F_{sw}$ from 50% to 62% on the same airfoil. The yellow lines indicate the shockwave positions. Moving the shockwave position rightward increases the chord distance from the suction peak to the shockwave's starting position, indirectly enlarging the area of the suction platform's fluctuation. Despite the absence of specific weights for the six pressure features in the model's conditional input, the actual results demonstrate varying degrees of importance among the features. The model tends to prioritize the suction peak, adjusting other pressure features accordingly after satisfying this criterion.

In airfoil design, reducing the strength of shockwaves decreases aerodynamic losses and drag, and helps improve the lift-to-drag ratio of airfoils (Okoronkwo et al. 2023). Fig. 16 (c) shows the inverse design results of the same airfoil with a gradual decrease in shockwave strength. These results have been verified by CFD, and the generated CP decreases according to the specified shockwave strength. Usually, the reduction of shockwave strength is achieved by reducing the height of the shockwave's starting position, resulting in a corresponding decrease in the area of suction platform fluctuation and an increase in the pressure gradient. As a key indicator of the recovery trend of the suction platform, changes in pressure gradient affect almost the entire airfoil upper surface features of the airfoil. As shown in Fig. 16(d), the change from negative to positive $F_{pg}$ is achieved by increasing the height of the suction peak and decreasing the height of the shockwave's starting position, which indirectly reduces the shockwave strength and the suction plateau fluctuation.

In transonic flow, the formation of dual shockwaves can lead to aerodynamic instability, increased drag, and higher thermal loads. Therefore, designers often focus on optimizing the airfoil geometry and flow field to minimize or avoid the formation of dual shockwaves. Fig. 16 (e) shows the inverse design results of avoiding dual shockwave formation by reducing the fluctuations of the suction platform. As the fluctuation decreases, the model jointly achieves this goal by adjusting the height of the shockwave's starting position to coincide with the height of the suction peak and moving the shockwave position to the left. This flattens the suction platform and helps control the rising trend of resistance in the transonic zone. Fig. 16 (f) presents the inverse design results of changing only the minimum pressure value on the lower surface. Notably, although the model generates a complete pressure distribution, it effectively distinguishes among the features of the upper and lower surfaces. It is clear that increasing only the minimum pressure value on the lower surface does not significantly affect the upper surface pressure features, indicating that our inverse design approach has learned some physical relationships and differentiates among various features.

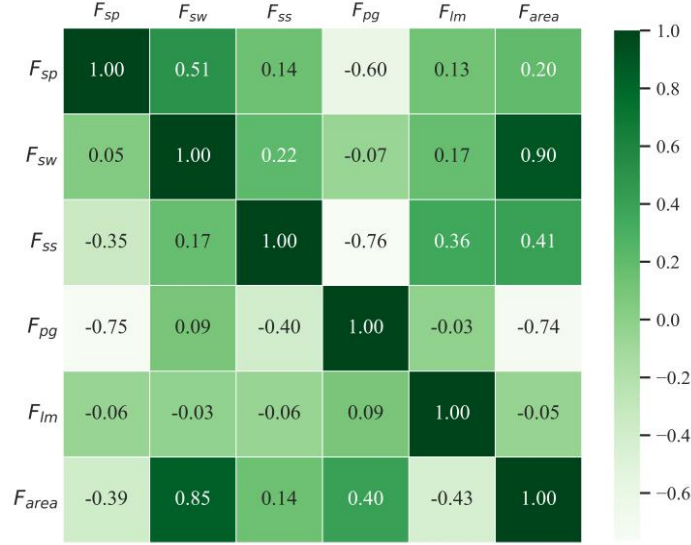

Fig. 17 Correlation among pressure features.

Despite the coupling and correlation between pressure features, the CDDPM-based inverse design framework can reveal the hidden patterns and provide valuable references for designers. Fig. 17 shows the correlation among the pressure features, where the vertical axis represents the pressure features that are changed, and the horizontal axis represents the affected pressure features. It can be seen from the figure that, for example, changing only $F_{sp}$ significantly affects $F_{sw}$ and $F_{pg}$, and the fluctuating change of $F_{pg}$ is mainly influenced by $F_{sp}$ and $F_{ss}$, which is in mutual agreement with the trend of Fig. 16. In summary, the proposed inverse design framework can accurately identify the significance of each pressure feature and design the corresponding airfoil geometry.

**5.3 Inverse optimization design based on pressure features**

To better meet the needs of designers in actual airfoil design and considering the significant influence of pressure features on airfoil performance, the design process not only needs to satisfy the constraints of these pressure features but also aims to maximize the lift-to-drag ratio (L/D) based on this foundation. Therefore, the optimization problem presented in Table 2 is defined as maximizing L/D of the airfoil under the constraints of the specified pressure features. Based on the EGO algorithm and combined with an active learning strategy, iterative loops are performed until the error constraints are satisfied, and the convergence history is shown in Fig. 18. Design 1 reaches the optimal solution for L/D after 81 iterations under the constraints of pressure features, where a small penalty value (50) is applied to the optimization results that do not satisfy these constraints. In Design 2, under unconstrained conditions and without the incorporation of penalty values, the optimal solution was attained after approximately ninety iterations.

**Table 2**

Optimization problem statement for inverse design.

|  | Function | Description |
| --- | --- | --- |
| Maximize | $L/D$ | Maximize the lift to drag ratio |
| Design variables | $F_{sp}, F_{sw}, F_{ss}, F_{pg}, F_{lm}, F_{area}$ | Specify pressure features |
| Constrains | $-1.5 < F_{sp} < -1.0$ | |
|  | $45\% < F_{sw} < 55\%$ | Design 1: Optimized airfoil with constraints |
|  | $F_{ss} < 0.7$ | Design 2: Optimized airfoil without constraints |
|  | $0 < F_{pg} < 0.2$ | |
|  | $F_{lm} > -0.35$ | |

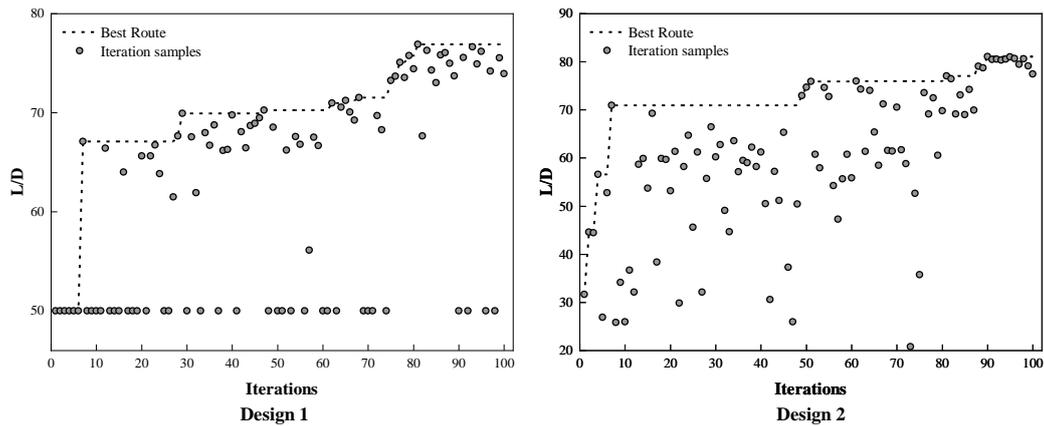

**Fig. 18** Convergence history of EGO-based optimization with active learning

The geometry of the airfoil is inverse designed based on the pressure features of the optimal L/D, and the final airfoils are shown in Fig. 19, where the objective function and pressure features values are summarized in Table 3. It can be seen that the inverse designed airfoil basically matches the CFD simulation results, and the L/D reaches 77.18 under the pressure feature constraints. It is worth noting that under unconstrained conditions, although the L/D achieves a better result, the fluctuating changes in the shock wave and the instability of the pressure gradient may render the airfoil unsuitable for practical design. If only the maximum L/D is pursued, the obtained airfoil may lack practical application value. In addition, compared with constrained optimization, unconstrained optimization has a more discrete parameter distribution and a slower convergence process. This may be due to the sampling space being larger than the dataset, resulting in more samples required for method training. However, for the preliminary design stage of the airfoil, the accuracy of the method is sufficient to support subsequent research. The method presented in this paper not only achieves the directional generation of airfoil geometries based on performance indicators, but also demonstrates better performance when constraints are applied. In addition, the method proposed in this paper provides a new approach for future efficient airfoil design and demonstrates a wide range of application prospects.

Table 3

Summary of pressure features and L/D for different designs

|  | L/D | $F_{sp}$ | $F_{sw}$ | $F_{ss}$ | $F_{pg}$ | $F_{lm}$ | $F_{area}$ |
|---|---|---|---|---|---|---|---|
|  | **Design 1**: Optimized airfoil with constraints | | | | | | |
| Optimum predicted | 76.91 | -1.41 | 0.52 | 0.62 | 0.14 | -0.09 | 0.0507 |
| Optimum calculated | 77.18 | -1.42 | 0.52 | 0.61 | 0.18 | -0.10 | 0.0445 |
|  | **Design 2**: Optimized airfoil without constraints | | | | | | |
| Optimum predicted | 81.05 | -1.37 | 0.59 | 0.49 | 0.36 | -0.21 | 0.0448 |
| Optimum calculated | 80.69 | 1.36 | 0.59 | 0.52 | 0.31 | -0.21 | 0.0532 |

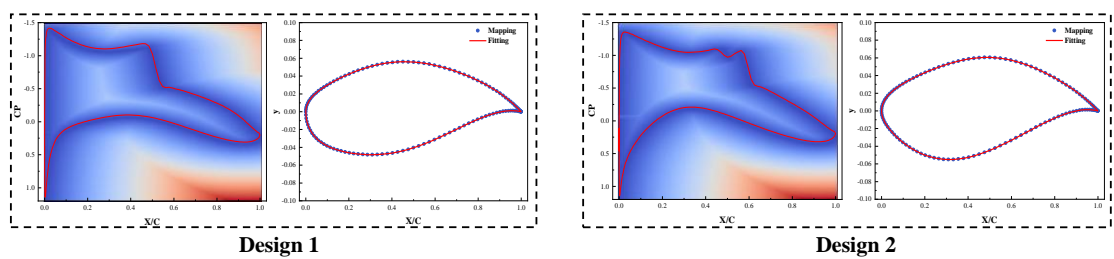

Design 1                                             Design 2

**Fig. 19** Pressure distribution and geometry of optimal airfoils

## 6 Conclusions

This paper proposes a highly precision and reliable airfoil inverse design method aimed at assisting designers in swiftly identifying suitable design directions and target airfoil shapes during the early stages of design. The method adopts CDDPM as the generative model to produce CP distributions corresponding to six specified pressure features. Subsequently, the CP distributions are converted into the required airfoil geometry by the mapping model. Through simulation experiments, the proposed method is validated to ensure both the accuracy of generated airfoil shapes and compliance with practical requirements in industrial design. Finally, the model framework is optimally updated using the EGO and combining active learning strategies.

Our research consolidates previous inverse design methodologies and proposes a generic inverse design process along with a dataset preparation procedure, laying the groundwork for future research endeavors. By introducing the diffusion model into airfoil inverse design, our approach surpasses existing research methods in design accuracy. Moreover, a practical method to readjust each performance indicator value is proposed, aiming to provide a rational combination of performance indicator values (i.e., meeting aerodynamic constraints) for the inverse design framework and addressing the challenging problem of selecting performance indicators that comply with aerodynamic constraints. Despite these advancements, our research still possesses certain limitations, and future work warrants further exploration in the following aspects.

1) To expedite the training process of the diffusion model. The training of the diffusion model is time-consuming, which to some extent limits the practicality and efficiency of the inverse design method. Future work could explore more effective training algorithms, optimize network structures, and other methods to accelerate the training process of the diffusion model.

2) Our current inverse design framework is implemented through three models, namely the diffusion model, the mapping model and the optimization model. In the future, it can be explored to integrate the three models into an end-to-end model on the premise of ensuring accurate extraction of aerodynamic features, thereby achieving a more efficient reverse design process.

**CRediT authorship contribution statement**

**Shisong Deng:** Conceptualization, Data curation, Writing – original draft, Writing – review & editing. **Qiang Zhang:** Funding acquisition, Projection administration, Supervision. **Zhengyang Cai:** Validation, Writing – review & editing.

**Acknowledgements**

This study is supported by the Young Scientists Fund of the National Natural Science Foundation of China (No. 72201087)

**Declarations**

**Conflict of interest** On behalf of all authors, the corresponding author states that there is no conflict of interest..

**Replication of Results** Code and data for replication can be provided up on request.